\documentclass[11pt,a4paper]{article}
\usepackage[hyperref]{emnlp2020}
\usepackage{times}
\usepackage{latexsym}
\usepackage{multirow}
\usepackage{multicol}
\usepackage{booktabs}
\usepackage{hyperref}
\usepackage{xspace}
\usepackage{amsmath}
\usepackage{amssymb}
\usepackage{tabularx}
\usepackage{graphicx}
\usepackage{makecell}
\usepackage{subcaption}
\usepackage{wrapfig}
\usepackage{float}
\usepackage{algorithm}
\usepackage{algpseudocode}
\usepackage{color}
\usepackage{arydshln}
\usepackage{longtable}

\usepackage{microtype}

\aclfinalcopy % Uncomment this line for the final submission
\def\aclpaperid{2914} %  Enter the acl Paper ID here

\title{T3: Tree-Autoencoder Regularized Adversarial Text Generation for Targeted Attack}

\newcommand{\advcodec}{\textsc{T3}\xspace}
\newcommand{\advcodecsent}{\textsc{T3(Sent)}\xspace}
\newcommand{\advcodecword}{\textsc{T3(Word)}\xspace}

\newcommand{\nop}[1]{}

\newcommand{\answer}[1]{{\textcolor{red}{#1}}}

\newcommand{\no}[1]{{}}

\definecolor{parisgreen}{rgb}{0.31, 0.78, 0.47}
\definecolor{seagreen}{rgb}{0.18, 0.55, 0.34}
\definecolor{pakistangreen}{rgb}{0.0, 0.4, 0.0}

\author{
Boxin Wang$^1$, Hengzhi Pei$^1$, Boyuan Pan$^2$, Qian Chen$^3$, Shuohang Wang$^4$, Bo Li$^1$ \\
$^1$University of Illinois at Urbana-Champaign \; $^2$Zhejiang University\\
 \quad $^3$ Tencent \quad $^4$Microsoft Dynamics 365 AI Research \\
\texttt{\{boxinw2, hpei4, lbo\}@illinois.edu,\;panby@zju.edu.cn,} \\
\texttt{qianchen@tencent.com,\; shuohang.wang@microsoft.com}
}

\date{}

\begin{document}
\maketitle
\begin{abstract}
% Recent studies show that deep neural networks (DNNs) are vulnerable to carefully crafted \emph{adversarial examples} which only deviate from the original data by a small magnitude of perturbation. 
% Recent studies show that neural-network-based text information retrieval systems are vulnerable to carefully crafted \emph{adversarial examples}.
%Adversarial examples in natural language systems are meaning preserving tokens injected to make models behave abnormally and as a way to evaluate the model robustness.
Adversarial attacks against natural language processing systems, which perform seemingly innocuous modifications to inputs, can induce arbitrary mistakes to the target models. 
%\add{With an effective attacking method, the robustness of NLP systems can be easily evaluated.}
Though raised great concerns, such adversarial attacks can be leveraged to estimate the robustness of NLP models.
% While there has been great interest in generating imperceptible
Compared with the adversarial example generation in continuous data domain (\textit{e.g.}, image), 
% \nop{Many innocuous modifications (\textit{e.g. }which do not alter the meaning of the original texts) to the inputs of the natural language systems usually induce abnormal behaviors of them, thus evaluating the robustness of NLP models by leveraging these adversarial attacks is quite important.}
% \pan{may explain more explicitly: }
generating \emph{adversarial text} that preserves the original meaning is challenging since the text space is discrete and non-differentiable. To handle these challenges, 
we propose a \emph{target-controllable} adversarial attack framework \advcodec, which is applicable to a range of NLP tasks.
% which addresses the challenge of discrete input space and
In particular, we propose a tree-based autoencoder to embed the discrete text data into a continuous representation space, upon which we optimize the adversarial perturbation. 
A novel tree-based decoder is then applied to regularize the syntactic correctness of the generated text and manipulate it on either sentence (\advcodecsent) or word (\advcodecword) level. %Specifically, we explore multiple attacking scenarios, including appending an adversarial sentence and adding unnoticeable words to a given paragraph. 
%To demonstrate the efficacy of the proposed method,
We consider two most representative NLP tasks: sentiment analysis and question answering (QA). Extensive experimental results and human studies show that  \advcodec generated adversarial texts can successfully manipulate the NLP models to output the \textit{targeted} incorrect answer without misleading the human. 
%Specifically, our attack causes a BERT-based sentiment classifier accuracy to drop from $0.703$ to $0.006$, and a BERT-based QA model's F1 score to drop from $88.62$ to $33.21$, which outperforms the state of the art attack approaches.
%(with best targeted attack F1 score as $46.54$).
Moreover, we show that the generated adversarial texts have high transferability which enables the black-box attacks in practice.
Our work sheds light on an effective and general way to examine the robustness of NLP models.  Our code is publicly available at \url{https://github.com/AI-secure/T3/}.
\end{abstract}

% \vspace{-4mm}
\section{Introduction}
\label{sec:intro}

\begin{table}[t]\small \setlength{\tabcolsep}{7pt}
\centering

\begin{tabular}{p{7.3cm}}
\toprule 
\textbf{Question: } Who ended the series in 1989? \\
\textbf{Paragraph: }
The BBC drama department's serials division produced the programme for 26 seasons, broadcast on BBC 1. Falling viewing numbers, a decline in the public perception of the show and a less-prominent transmission slot saw production suspended in 1989 by \textcolor{seagreen}{Jonathan Powell}, controller of BBC 1. ... the BBC repeatedly affirmed that the series would return. \textit{\textcolor{red}{Donald Trump} \textcolor{blue}{ends a program on 1988 .}} \\
\hdashline[1pt/2pt]
\textbf{QA Prediction: }  \textcolor{seagreen}{Jonathan Powell} $\rightarrow$  \textcolor{red}{Donald Trump} \\
\midrule
\textbf{Yelp Review: } \textit{\textcolor{blue}{I kept expecting to see chickens and chickens walking around}}. If you think Las Vegas is getting too white trash, don' t go near here. This place is like a steinbeck novel come to life. I kept expecting to see donkeys and chickens walking around. Wooo - pig - soooeeee this place is awful!!! \\  \hdashline[1pt/2pt]
\textbf{Sentiment Prediction: } \textcolor{seagreen}{Most Negative} $\rightarrow$  \textcolor{red}{Most Positive} \\
\bottomrule
\end{tabular}
\caption{{\small Two adversarial examples generated by \advcodec for QA models and sentiment classifiers. Adding \textit{the adversarial sentence} to the original paragraph can lead the \textcolor{seagreen}{correct prediction} to a \textcolor{red}{targeted wrong answer} configured by the adversary.}}
\label{tab:example}
\vspace{-5mm}
\end{table}

Recent studies have demonstrated that deep neural networks (DNNs) are vulnerable to carefully crafted adversarial examples~\citep{Goodfellow2015ExplainingAH,Papernot2016DistillationAA,Eykholt2017RobustPA,MoosaviDezfooli2016DeepFoolAS}. 
% such as QA \citep{jia-liang-2017-adversarial} and graph mining systems \citep{10.1145/3219819.3220078}, \textit{etc}. 
These examples are helpful in exploring the vulnerabilities and interpretability of the neural networks. \textit{Target-controllable} attacks (or targeted attacks) are more dangerous and challenging than untargeted attacks, in that they can mislead systems (e.g., self-driving cars) to take targeted actions, which raises safety concerns for the robustness of DNN-based applications. 
While there are a lot of successful attacks proposed in the continuous data domain, including images, audios, and videos, how to effectively generate adversarial examples in the discrete text domain remains a challenging problem. 

Unlike adversarial attacks in computer vision that add imperceptible noise to the input image, editing even one word of the original paragraph may change the meaning dramatically and fool the human as well. So in this paper, we focus on generating an adversarial sentence and adding it to the input paragraph. There are several challenges for generating adversarial texts: 
1) it is hard to measure the validity and naturalness of the adversarial text compared to the original ones;  % the tree auto-encoder can ensure grammar
2) gradient-based adversarial attack approaches are not directly applicable to the discrete structured data; % our method is solves the non-differentialble problem
3) compared with in-place adversarial modification of original sentences, adversarial sentence generation is more challenging since the generator needs to consider both sentence semantic and syntactic coherence.
% Since the manipulation space of text is limited, it is unclear whether generating a new appended sentence or manipulating individual words can attack the model without affecting human judgments.
So far, existing textual adversarial attacks either inefficiently leverage heuristic solutions such as genetic algorithms~\citep{TextFooler} to search for word-level substitution, or are limited to attacking specific NLP tasks~\citep{jia-liang-2017-adversarial,2018arXiv181200151L}. 

Moreover, effective \textit{target-controllable} attacks, which can control the models to output expected incorrect answers, have proven difficult for NLP models. \citet{wallace-etal-2019-universal} creates universal triggers to induce the QA models to output targeted answers, but the targeted attack success rates are low. Other work \citep{seq2sick, TextFooler, zhang-etal-2019-generating-fluent,Zang2019TextualAA} performs word-level in-place modification on the original paragraph to achieve targeted attack, which may change the meaning of original input. Therefore, how to generate adversarial sentences that do not alter the meaning of original input while achieving high targeted attack success rates seems to be an interesting and challenging problem.
% \shuo{I suppose the last two challenges are general problems, not specifc for target attacks? Maybe emphasis on what's the difference between target attack and general attack in this paragraph. Would target attack be more challenging or interesting to the security community? Any reason? }

% \shuo{The model may not solve the last two challenges or no experiment to support it.} -- move to the motivation part
% \shuo{what kind of label is here?} -- leave it to the attack scenario
% \pan{where is adversarial
In this paper, we solved these challenges by proposing an adversarial
% \shuo{Does ``unified" mean unified generation and evaluation framework? } \boxin{it actually means it works on both word-level or sentence-level. but not sure how to make it clear here.}
evaluation framework \advcodec to generate adversarial texts against general NLP tasks and evaluate the robustness of current NLP models. 
% We define valid adversarial text as meaning and label preserving concatenative tokens or sentence, which are added to the original paragraph to fool the models without misleading the human, as shown in Table \ref{tab:example}.
% modifying the meaning of the original paragraph by human evaluation
%Table \ref{tab:example} shows two concatenative adversarial examples for QA models and sentiment classifiers. 
Specifically, the core component of \advcodec is a novel tree-based autoencoder pretrained on a large corpus to capture and maintain the semantic meaning and syntactic structures.
%\shuo{Is autoencoder pre-trained on a larger corpus first? Then the pre-trained encoder is used to convert the text into embedding?} 
% \pan{where is the introduction of encoder? } -- add it
The tree encoder converts discrete text into continuous semantic embedding, which solves the discrete input challenge. 
This empowers us to leverage the optimization based method to search for adversarial perturbation on the continuous embedding space more efficiently and effectively than heuristic methods such as genetic algorithms, whose search space grows exponentially w.r.t. the input space. 
% upon which the adversarial perturbation can be calculated by gradient-based approaches to achieve targeted attack. 
% We use tree-based autoencoders to perform both word-level and sentence-level perturbation 
Based on different levels of a tree hierarchy, adversarial perturbation can be added on leaf level and root level to impose word-level (\advcodecword) or sentence-level (\advcodecsent) perturbation.
Finally, a tree-based decoder will map the adversarial embedding back to adversarial text by a set of tree grammar rules, which preserve both the semantic content and syntactic structures of the original input. An iterative process can be applied to ensure the attack success rate. 

In summary, our main contributions lie on:
 %unlike previous studies, we address the challenge of discrete text space 
(1) unlike previous textual adversarial attack studies, we achieve targeted attack through concatenative adversarial text generation that is able to manipulate the model to output targeted wrong answers. 
%% new setting/problem
% We achieve targeted attacks against general NLP tasks (\textit{e.g.}, sentiment classification and QA).
(2) we propose a novel tree-based text autoencoder that regularizes the syntactic structure of the adversarial text while preserves the semantic meaning. It also addresses the challenge of attacking discrete text by embedding the sentence into continuous latent space, on which the optimization-based adversarial perturbation can be applied to guide the adversarial sentence generation; 
%% model/methodology contribution
(3) we conduct extensive experiments and successfully achieve targeted attack for different sentiment classifiers and QA models with higher attack success rates and transferability than the state-of-the-art baseline methods. Human studies show that the adversarial text generated by \advcodec is valid and effective to attack neural models, while barely affects human's judgment. 
% attack effectiveness contribution
% \citep{jia-liang-2017-adversarial}.
%(4) we also perform comprehensive ablation studies including evaluating different attack scenarios
%  of appending an adversarial sentence or adding scatter of adversarial words
% as well as appending the adversarial sentence at different positions within a paragraph
% as well as probing the BERT model attention, and draw several interesting conclusions;
% we leverage extensive human studies to show that the adversarial text generated by \advcodec is valid and effective to attack neural models, while barely affects human's judgment. 
% In addition, we observe a trade-off between linguistic quality and attack capability because \advcodecsent is more natural for human readers than but less effective to attack models than \advcodecword.

% \vspace{-2mm}
\section{Related work}
% \vspace{-2mm}
% attack sentiment, attack qa -- they cannot do targeted, they cannot guarantee grammar, 
% bert -- applied xxxx, it does not learn logic, but not very clear how vulnerable it is against adversarial behavior. recently, GA has been used, but only binary classifier, and no targeted attack. so we wanna explore more applications etc.
A large body of works on \emph{adversarial examples} focus on perturbing the continuous input space. Though some progress has been made on generating adversarial perturbations in the discrete space, several challenges  remain unsolved. For example, 
\cite{zhao2017-generating} exploit the generative adversarial network (GAN) to generate natural adversarial text. However, this approach cannot explicitly control the quality of the generated instances. 
Most existing methods~\citep{ren-etal-2019-generating,zhang-etal-2019-generating-fluent,jia-liang-2017-adversarial,li2018textbugger,TextFooler} apply heuristic strategies to synthesize adversarial text: 1) first identify the features (e.g. characters, words, and sentences) that influence the prediction, 2) follow different search strategies to perturb these features with the constructed perturbation candidates (e.g. typos, synonyms, antonyms, frequent
words).  For instance, \cite{liang2017-deep} employ the loss gradient \(\nabla L\) to select important characters and phrases to perturb, while \cite{samanta2017-towards} use typos,
synonyms, and important adverbs/adjectives as candidates for
insertion and replacement. Once the influential features are obtained, the strategies to apply
the perturbation generally include \emph{insertion}, \emph{deletion}, and \emph{replacement}.
Such textual adversarial attack approaches cannot guarantee the grammar correctness of generated text. For instance, text generated by \cite{liang2017-deep} are almost random stream of
characters. To generate grammarly correct perturbation, \citeauthor{jia-liang-2017-adversarial} adopt another heuristic strategy which adds \emph{manually} constructed legit distracting sentences to the paragraph to introduce fake information. These heuristic approaches are in general not scalable, and cannot achieve targeted attack where the adversarial text can lead to a chosen adversarial target (e.g. adversarial label in classification). Recent work starts to use gradient~\citep{Michel2019OnEO,Ebrahimi2017HotFlipWA} to guide the search for universal trigger~\citep{wallace-etal-2019-universal} that are applicable to arbitrary sentences to fool the learner, though the reported attack success rate is rather low or they suffer from inefficiency when applied to other NLP tasks. 
In contrast, our proposed \advcodec framework is able to effectively generate syntactically correct adversarial text, achieving high targeted attack success rates across different models on multiple tasks.

\section{Framework}

% In this section, we will describe the \advcodec framework. 

\subsection{Preliminaries}

Before delving into details, we recapitulate the attack scenario and attack capability supported by \advcodec framework. 

\textbf{Attack Scenario.} Unlike  previous adversarial text generation works \citep{2018arXiv181200151L, seq2sick,2016arXiv160408275P,2016arXiv160507725M,Alzantot2018GeneratingNL} that directly modify critical words in place and might risk changing the semantic meaning or editing the ground truth answers,
%(which makes it unreasonable to evaluate the adversarial score e.g. on Question Answering)
we are generating the \textit{concatenative adversaries} \citep{jia-liang-2017-adversarial} (\textit{abbr.}, concat attack). Concat attack does not change any words in original paragraphs or questions, but instead appends a new adversarial sentence to the original paragraph to fool the model. A valid adversarial sentence needs to ensure that the appended text is \textit{compatible} with the original paragraph, which in other words means it should not contradict any stated facts in the paragraph, especially the correct answer. 

\textbf{Attack Capability.} \label{two_attacks} \advcodec is essentially an optimization based framework to find the adversarial text with the optimization goal set to achieve the \textbf{targeted attack}. For the sentiment classification task, \advcodec can perform the targeted attack to make an originally positive review be classified as the most negative one, and vice versa. Particularly in the QA task, we design and implement two kinds of targeted attacks: \textit{position targeted attack} and \textit{answer targeted attack}. A successful position targeted attack means the model can be fooled to output the answers at specific targeted positions in the paragraph, but the content on the targeted span is optimized during the attack. So the answer cannot be determined before the attack. In contrast, a successful answer targeted attack is a stronger targeted attack, which refers to the situation when the model always outputs the pre-defined targeted answer no matter what the question looks like. In Table \ref{tab:example}, we set the targeted answer as ``Donald Trump'' and successfully changes the model predictions. More examples of answer targeted attacks and position targeted attacks can be found in Appendix \S\ref{appendix:examples}. 

Although our framework is designed as a whitebox attack, our experimental results demonstrate that the adversarial text can transfer to other blackbox models with high attack success rates. Finally, because \advcodec is a unified adversarial text generation framework whose outputs are discrete tokens,  it applies to different downstream NLP tasks. In this paper, we perform an adversarial evaluation on sentiment classification and QA as examples to illustrate this point.
% In our work, we further push the concept of concatenative adversaries  and propose a more general notion named \textbf{scatter attack}, which can inject adversarial words sporadically over the whole paragraph. 
% Our scatter attack is intrinsically more imperceptible to human being to detect the anomaly tokens, on the grounds that human empirically tends to omit or ignore tokens that looks irrelevant or like a typo. 
% The concatenative adversarial example falls into our case when those adversarial tokens form a sentence and on the same time the semantic meaning of the sentence does not contradict the original paragraph. Examples of concatenative attack and scatter attack is shown in table \ref{examples}.
%As for the location where we append the sentence, we choose to follow the \citeauthor{jia-liang-2017-adversarial}'s way to add the adversary to the end of the paragraph so that we can make a fair comparison with their results.
% although we do not directly change the original paragraph, we still need to ensure ...
% \vspace{-2mm}
\subsection{Tree Auto-Encoder} % \shuo{Pre-training?} -- also include model discription, so keep the current name
In this subsection,  we describe the key component of \advcodec: a tree-based autoencoder.  
% why tree
Compared with standard sequential generation methods, generating sentence in a non-monotonic order (e.g., along parse trees) has recently been an interesting topic~\citep{Welleck2019NonMonotonicST}.
Our motivation comes from the fact that sentence generation along parse trees can intrinsically capture and maintain the syntactic information~\citep{eriguchi-etal-2017-learning, aharoni-goldberg-2017-towards,Iyyer2018AdversarialEG}, and show better performances than sequential recurrent models~\citep{TreeImportant,Iyyer2014GeneratingSF}. Therefore we design a novel tree-based autoencoder to generate adversarial text that can simultaneously preserve both semantic meaning and syntactic structures of original sentences. Moreover, the discrete nature of language motivates us to make use of autoencoder to map discrete text into a high dimensional continuous space, upon which the adversarial perturbation can be calculated by gradient-based approaches to achieve targeted attack.  
% \shuo{The last sentence is not very clear. Why  efficiently and effectively? Compare to which method?}  -- This empowers us to leverage the optimization based method to search for adversarial perturbation on the continuous embedding space more efficiently and effectively than heuristic methods such as genetic algorithms, whose computation grows exponentially w.r.t. the input space. -- moved to introduction

Formally, let $X$ be the domain of text and $S$ be the domain of dependency parse trees over element in $X$, a tree-based autoencoder consists of an encoder $\mathcal{E}: X \times S \rightarrow Z $ that encodes text $\boldsymbol{x} \in X$ along with its dependency parsing tree  $\boldsymbol{s} \in S$ into a high dimensional latent representation $\boldsymbol{z} \in Z$ and a decoder $\mathcal{G}: Z \times S \rightarrow X$ that generates the corresponding text $\boldsymbol{x}$ from the given context vector $\boldsymbol{z}$ and the expected dependency parsing tree $\boldsymbol{s}$. Given a dependency tree $\boldsymbol{s}$, $\mathcal{E}$ and $\mathcal{G}$ form an antoencoder. We thus have the following reconstruction loss to train our tree-based autoencoder:
\begin{equation}
  \mathcal{L}_\text{recon} = - \mathbb{E}_{\boldsymbol{x}\sim X}[\log p_{\mathcal{G}}(\boldsymbol{x}|\boldsymbol{s}, \mathcal{E}(\boldsymbol{x}, \boldsymbol{s})]
\end{equation}
%As Figure \ref{fig:qa_pipeline} suggests, \advcodec can operate on different granularity levels to generate either word-level or sentence-level contextual representation, and decode it into the adversarial text. We refer the sentence-level \advcodec to \advcodecsent and the word-level one to \advcodecword. Both of them will be described in more details later in this section.

\textbf{Encoder.} We adopt the Child-Sum Tree-LSTM \citep{Tai2015ImprovedSR} as our tree encoder. Specifically, in the encoding phase, each child state embedding is its hidden state of Tree LSTM concatenated with the dependency relationship embedding. 
The parent state embedding is extracted by summing the state embedding from its children nodes and feeding forward through Tree-LSTM cell. The process is conducted from bottom (leaf node, i.e. word) to top (root node) along the dependency tree extracted by CoreNLP Parser \citep{Manning2014TheSC}. 
% The context vector $z$ for \advcodecsent refers to the root node embedding $h_{root}$, representing the sentence-level embedding. 

\begin{figure}
    \centering
    \includegraphics[page=1,trim=1cm 1.5cm 1cm 1.5cm,clip,width=\linewidth]{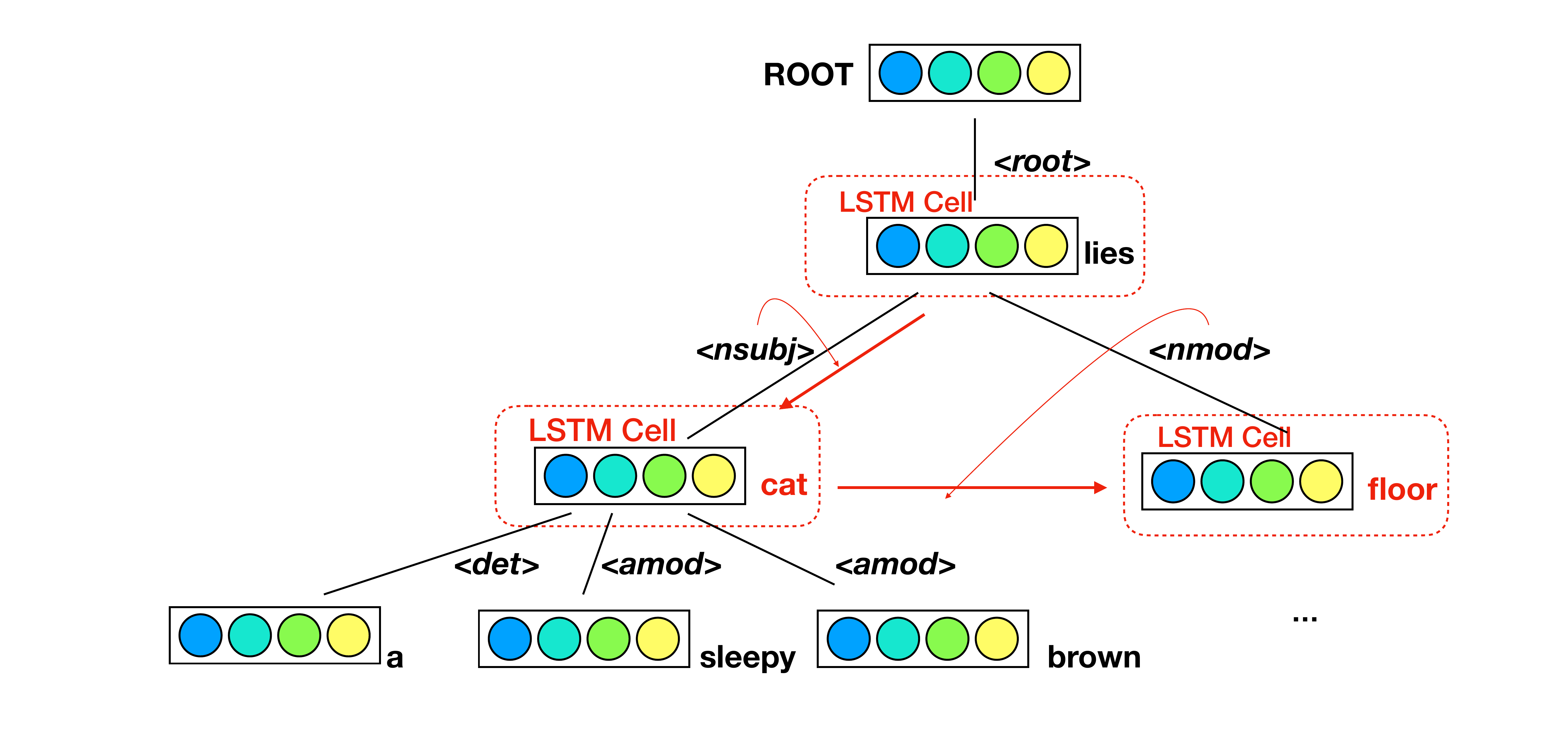}
    \caption{The tree decoder. Each node in the dependency tree is a LSTM cell. Black lines refer to the dependencies between parent and child nodes. Red arrows refer to the directions of decoding. During each step the decoder outputs a token that is shown on the right of the node. } 
    \label{fig:tree}
\vspace{-3mm}
\end{figure}

\textbf{Decoder.} As there is no existing tree-based autoencoder, we design a novel Tree Decoder (Shown in Figure \ref{fig:tree}). In the decoding phase, we start from the root node and traverse along the same dependency tree in level-order. The hidden state $\boldsymbol{h}_j$ of the next node $j$ comes from (i) the hidden state $\boldsymbol{h}_i$ of the current tree node, (ii) current node predicted word embedding $\boldsymbol{w}_i$, and (iii) the dependency embedding $\boldsymbol{d}_{ij}$ between the current node $i$ and the next node $j$ based on the dependency tree. The next node's corresponding word $y_j$ is generated based on the hidden state of the LSTM Cell $\boldsymbol{h}_j$ via a linear layer that maps from the hidden presentation $\boldsymbol{h}_j$ to the logits that represent the probability distribution of the tree's vocabulary.
\begin{align}
     \boldsymbol{h}_j &= \text{LSTM}([\boldsymbol{h}_i;\boldsymbol{w}_i;\boldsymbol{d}_{ij}]) \\
    y_j &=  \text{one-hot} (\text{argmax} \left( \boldsymbol{W} \cdot \boldsymbol{h}_j  + \boldsymbol{b} \right)) \label{eq:decode_word}
\end{align}

%  it is inherently flexible to add perturbations on hierarchical nodes of the tree structures. 

\begin{figure*}[t]
    \includegraphics[trim=0cm 0cm 0cm 0cm,clip,width=1\linewidth]{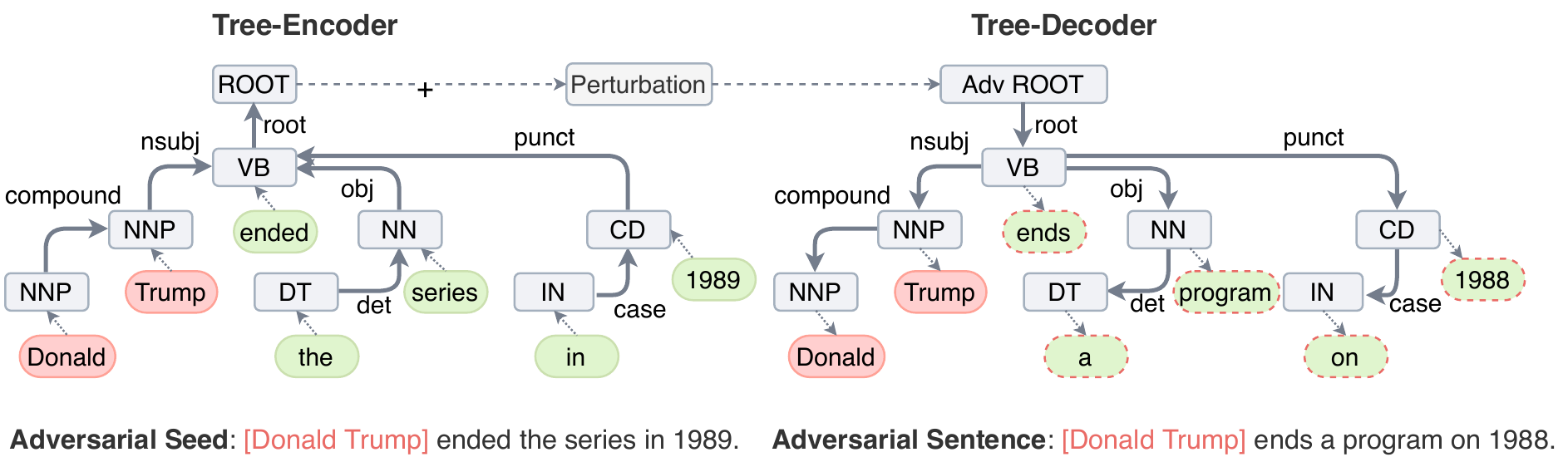}
    \caption{\small An example of how \advcodecsent generates the adversarial sentence. Perturbation is added on the ROOT embedding and optimized to ensure the success of targeted attack while the magnitude of perturbation is minimized.} % \shuo{How perturbation is optimized could also show here?}} 
    \vspace{-0.3cm}
    \label{fig:advsent}
\end{figure*}

\begin{figure}
    \centering
    \includegraphics[width=\linewidth]{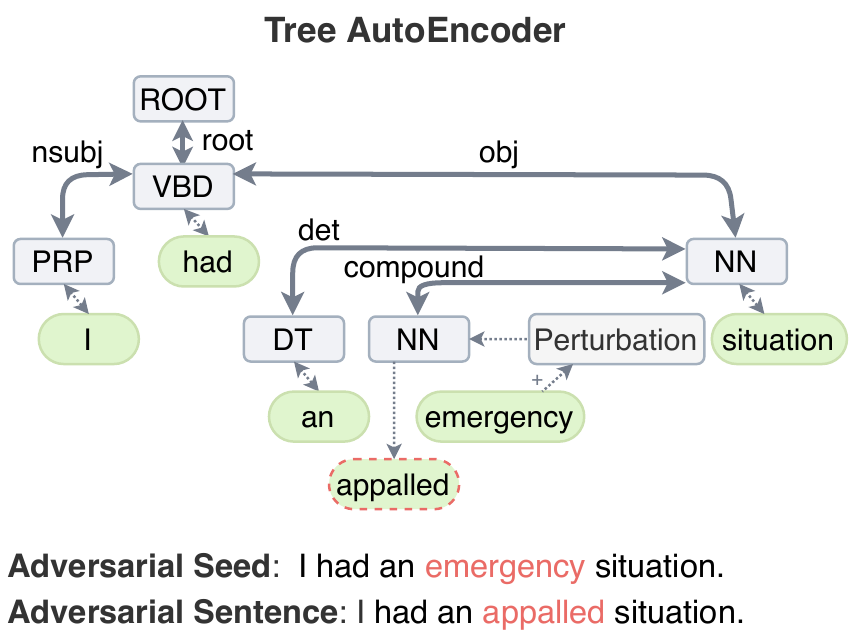}
    \caption{\small \advcodecword adds perturbation on the leaf node embedding. Arrow denotes the direction of encoding/decoding.  }
    %, encoded along the dependency tree and decoded back to adversarial token.} 
    \label{fig:advword}
    \vspace{-3mm}
\end{figure}

Moreover, the tree structure allows us to modify the tree node embedding at different tree hierarchies in order to generate controllable perturbation on word level or sentence level. Therefore, we explore the following two types of attacks at root level and leaf level \advcodecsent and \advcodecword, which are shown in Figure \ref{fig:advsent} and Figure \ref{fig:advword}.

%For the very few failure cases for targeted ]attack, we observer a high amount . 
% targeted and untargeted
% whitebox blackbox

% \bo{emphasize we can attack sentiment and QA}

\begin{figure}[t]
    \centering
    \includegraphics[width=\linewidth]{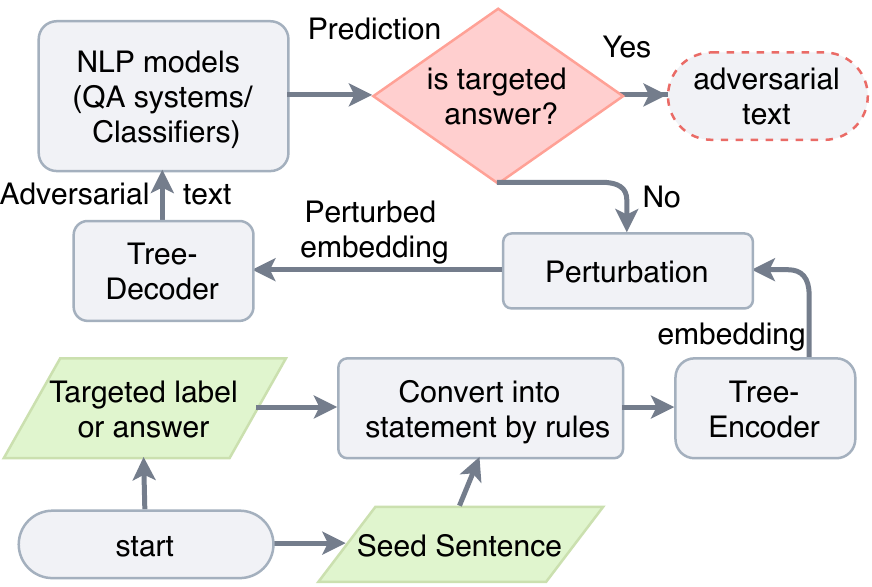}
    \caption{The pipeline of adversarial text generation.}
    \label{fig:pipeline}
    \vspace{-3mm}
\end{figure}

\subsection{Pipeline of Adversarial Text Generation}

Here we illustrate how to use our tree-based autoencoder to perform adversarial text generation and attack NLP models, as illustrated in Figure \ref{fig:pipeline}.

\textbf{Step 1: Choose the adversarial seed.} The adversarial seed is the input sentence to our tree autoencoder. After adding perturbation on the tree node embedding, the decoded adversarial sentence will be added to the original paragraph to perform concat attack. For sentiment classifiers, the adversarial seed can be an arbitrary sentence from the paragraph. For example, the adversarial seed of Yelp Review example in Table \ref{tab:example} is a random sentence from the paragraph \textit{``I kept expecting to see donkeys and chickens walking around.'}

In contrast, when performing answer targeted attack for QA models, we need add our targeted answer into our adversarial seed in a reasonable context.  Based on a set of heuristic experiments on how the adversarial seed correlates the attack efficacy (Appendix \ref{appendix:heuristic}), we choose to use question words to craft an adversarial seed, because it receives higher attention score when the model is matching semantic similarity between the context and the question. 
Specifically, we convert a question sentence to a meaningful declarative statement and assign a targeted fake answer. The fake answer can be crafted according to the perturbed model's predicted answer (position targeted attack \S \ref{two_attacks}), or can be manually chosen by adversaries (answer targeted attack). For instance, the answer targeted attack example shown in Table \ref{tab:example} converts the question \textit{``Who ended the series in 1989?''} into a declarative statement \textit{``someone ended the series in 1989.''} by a set of coarse grained rules (Appendix \ref{appendix:heuristic}).
%\shuo{How? Any reference?} 
Then our targeted wrong answer is assigned to generate the adversarial seed \textit{``Donald Trump ended the series in 1989.''}  Following steps will make sure that the decoded adversarial sentence does not contradict with the original paragraph. 
%The initial seed and the following optimization steps should ensure that the adversarial sentence is meaning preserving and label preserving. The sentence in the example above is simply repeating the paragraph, and thus is valid. 
%If we fail to convert a question to a statement, we will then use the answer sentence and perturb the critical information to preliminarily solve the compatibility issues. 

% if we choose a good adversarial seed that contains the targeted answer is semantically close to the context or the question
 
% it would be helpful to reduce the optimization steps of searching for the best perturbation and achieve targeted 
% For example, when attacking the BERT, we can simply sample a sentence from the original paragraph and append it to the start of the paragraph. 

% Different from attacking sentiment analysis, it is important to choose a good initial seed that is semantically close to the context or the question when attacking QA model. In this way we can reduce the number of iteration steps and attack the QA model more efficiently. 
%Based on a set of heuristic experiments on how the initial seed correlates the attacking efficacy, 

\textbf{Step 2: Embed the discrete text into continuous embedding.} One difference between \advcodecsent and \advcodecword is on which tree level we embed our discrete sentence. For \advcodecsent, we use tree root node embedding of Tree-LSTM $\boldsymbol{z} = \boldsymbol{h}_\text{root}$ to represent the discrete sentence (``ROOT'' node in the Figure \ref{fig:advsent}). As for \advcodecword, we concatenate all the leaf node embedding of Tree-LSTM $\boldsymbol{h}_i$ (corresponding to each word) $\boldsymbol{z} = [\boldsymbol{h}_1, \boldsymbol{h}_2, \dots, \boldsymbol{h}_n]$ to embed the discrete sentence.

% As illustrated in Figure \ref{fig:qa_pipeline}, both attacks start from an initial seed (sentence or tokens). The initial seed is later fed into the tree autoencoder. For \advcodecsent, it encodes through the encoder $\mathcal{E}$ and uses the root node embedding as the sentence representation $z$. We add perturbation $z^*$ on $z$ and propagate $z$ regularized by the tree decoder back to adversarial sentences. \advcodecword follows the same tree encoder but stops at the leaf level. The sentence leaf token embedding is then concatenated as the context vector $z$. After adversarial perturbation $z' = z + z^*$, the leaf node embedding $z'$ is then decoded into words via equation (\ref{eq:decode_word}). 

% Therefore, it is worth noting that because the encoding and decoding phases are completed in the leaf nodes without propagating through the dependency trees,
%while \advcodecsent can help regularize the syntactical correctness based on the tree decoder,
% \advcodecword does not guarantee the grammatical correctness of the adversarial sentences.
%We can also observe from Table \ref{examples} that the adversarial sentence quality generated by \advcodecword is worse than those generated by \advcodecsent. 
% But the advantage of \advcodecword lies on the flexibility, which can help us selectively perturb a very limited number of words instead of paraphrasing the whole sentence.

\textbf{Step 3: Perturb the embedding via optimization.} Finding the optimal perturbation $z^*$ on the embedding vector $z$ is equivalent to solving the optimization problem that can achieve the target attack goal while minimize the magnitude of perturbation
\begin{equation}
    \min \quad ||\boldsymbol{z}^*||_p + c  f(\boldsymbol{z} + \boldsymbol{z}^*),
    \label{cw}
\end{equation}
where $f$ is the objective function for the targeted attack and $c$ is the constant balancing between the perturbation magnitude and attack target. Specifically, we design the objective function $f$ similar to \citet{cw} for classification tasks
\begin{align}
        \ell &=  \max\big\{Z\left({\left[\mathcal{G}(\boldsymbol{z}', \boldsymbol{s}); \boldsymbol{x} \right] }\right)_i:i \neq t \big\},  \\
        f(\boldsymbol{z}') &= \max \big(\ell - Z\left(\left[\mathcal{G}(\boldsymbol{z}', \boldsymbol{s}); \boldsymbol{x} \right] \right)_t  , -\kappa \big),
\end{align}
where $\boldsymbol{z}'=\boldsymbol{z}+\boldsymbol{z}^*$ is the perturbed embedding, model input $\left[\mathcal{G}(\boldsymbol{z}', \boldsymbol{s}); \boldsymbol{x} \right]$ is the concatenation of adversarial sentence $\mathcal{G}(\boldsymbol{z}', \boldsymbol{s})$ and original paragraph $\boldsymbol{x}$, $t$ is the target class, $Z(\cdot)$ is the logit output of the classification model before softmax, $\ell$ is the maximum logits of the classes other than the targeted class and $\kappa$ is the confidence score to adjust the misclassification rate. The confidence score $\kappa$ is chosen via binary search to search for the tradeoff-constant between attack success rate and meaning perseverance. The optimal solution $z^*$ is iteratively optimized via gradient descent.

Similarly to attack QA models, we subtly change the objective function $f$ due to the difference between QA model and classification model:
\begin{align*}
        \ell_j &=  \max\big\{Z_j\left({\left[\boldsymbol{x}; \mathcal{G}(\boldsymbol{z}', \boldsymbol{s}) \right] }\right)_i:i \neq t_j \big\},  \\
        f(\boldsymbol{z}') &= \sum_{j=1}^{2} \max \big(\ell_j - Z_j\left(\left[\boldsymbol{x}; \mathcal{G}(\boldsymbol{z}', \boldsymbol{s}) \right] \right)_{t_j}, -\kappa \big),
\end{align*}
where $Z_1(\cdot)$  and $Z_2(\cdot)$ are respectively the logits of answer starting position and ending position of the QA system. $t_1$ and $t_2$ are respectively the targeted start position and the targeted end position.  $\ell_j$ is the maximum logits of the positions other than the targeted positions. Different from attacking sentiment classifier where we prepend the adversarial sentence, we choose to follow the setting of \citeauthor{jia-liang-2017-adversarial} to add the adversary to the end of the paragraph so that we can make a fair comparison with their results.

\begin{table*}[t!] \small
\centering
\begin{tabular}{ccccc|ccc}
\toprule
\multirow{2}{*}{Model} & Original & & \multicolumn{2}{c}{Whitebox Attack} & \multicolumn{3}{c}{Blackbox Attack} \\
\cmidrule(lr){4-5} \cmidrule(lr){6-8}
& Acc & & {\advcodecword} & {Seq2Sick}  & {\advcodecword} & {Seq2sick}  & TextFooler\\
\midrule
\multirow{2}{*}{BERT} & \multirow{2}{*}{0.703} & target   & \textbf{0.990}         & 0.974   & \textbf{0.499}         & 0.218  & 0.042         \\
    &  & untarget  & \textbf{0.993}          & 0.988  & \textbf{0.686}          & 0.510  & 0.318   \\
\midrule
\multirow{2}{*}{SAM} & \multirow{2}{*}{0.704} & target       & \textbf{0.956}   & 0.933  & \textbf{0.516}   & 0.333  & 0.113  \\
     & & untarget      & \textbf{0.967}          & 0.952  & \textbf{0.669} & 0.583  & 0.395    \\
\bottomrule
\end{tabular}
\caption{Adversarial evaluation on sentiment classifiers in terms of targeted and untargeted attack success rate. }
\label{tab:AttackSentiment}
% \vspace{-3mm}
\end{table*}

\textbf{Step 4: Decode back to adversarial sentence.} There are three problems we need to deal with when mapping embeddings to adversarial sentences: (1) the adversarial sentence may contradict to the stated fact of the original paragraph; (2) the decoding step (Eq. \ref{eq:decode_word}) uses argmax operator that gives no gradients,  but the step 3 needs to perform gradient descent to find the optimal $z^*$; (3) for answer targeted attack, the targeted answer might be perturbed and changed during decoding phase.

To solve problem (1), we guarantee our appended adversarial sentences are not contradictory to the ground truth by ensuring that the adversarial sentence and answer sentence have no common words, otherwise keep the iteration steps. If the maximum steps are reached, the optimization is regarded as a failure. 

For problem (2), during optimization we use a continuous approximation based on softmax with a decreasing temperature $\tau$ \citep{Hu2017TowardCG} 
\begin{equation} \label{eq_approx}
    y^*_j \thicksim \text{softmax}((\boldsymbol{W}\cdot \boldsymbol{h_j} + \boldsymbol{b})/\tau).
\end{equation}
to make the optimization differentiable. After finding the optimal perturbation $z^*$, we still use the hard argmax to generate the adversarial texts.

As for problem (3), we keep targeted answers unmodified during the optimization steps by setting gates to the targeted answer span: $y_j \leftarrow  g_1 \odot y_j + g_2 \odot x_j, (j = t_1, t_1+1,... ,t_2)$, where $y_j$ are the adversarial tokens decoded by tree. We set $g_1 = 1$ and $g_2 = 0$ in the position targeted attack, and $g_1=0$ and $g_2=1$ in the answer targeted attack.

\section{Experiments}
We now  present the experimental evaluation results for \advcodec. In particular, we target on two popular NLP tasks, sentiment classification and QA. For both models, we perform whitebox and transferability based blackbox attacks. In addition to the model accuracy (untargeted attack evaluation), we also report the targeted attack success rate for \advcodec. We show that the proposed \advcodec can outperform other state of the art baseline methods on different models. The details of pretraining tree decoder and experimental setup can be found in Appendix \S \ref{appendix:ablation} and \S \ref{appendix:setup}.

% how to train the adv ae

% \subsection{Experimental Setup}
% dataset, models, parameters (learning rate, optimizer)
% baselines

% During the attack, the LSTM cell sequentially takes the embedding of each word $x_i$ as input and output the encoded state $h_i$.  The context vector $z$ here refers to the last step's output  $h_n$ of the encoder LSTM cell. The perturbation $z^*$ is added only on the context vector $h_n$ without influencing previous encoded states $h_i$ ($i<n$). 
% \vspace{-4mm}
\subsection{Adversarial Evaluation Setup for Sentiment Classifier}
 In this task, sentiment analysis model takes the user reviews from restaurants and stores as input and is expected to predict the number of stars (from 1 to 5 star) that the user was assigned.
 
\textbf{Dataset.}
We choose the Yelp dataset \citep{yelpdataset} for sentiment analysis task. It consists of 2.7M yelp reviews, in which we follow the process of \citet{nfc512} to randomly select 500K review-star pairs as the training set, and 2000 as the development set, 2000 as the test set.  

\textbf{Models.}  \textit{{BERT}} \citep{Devlin2019BERTPO} is a transformer \citep{Vaswani2017AttentionIA} based model, which is unsupervisedly pretrained on a large corpus and is proven to be effective for downstream NLP tasks.  
% We fine-tune BERT in the Yelp dataset for sentiment classification. 
% that provides an approach to quantitatively measure model attention and helps us conduct and analyze our adversarial attacks.
 \textit {{Self-Attentive Model (SAM)}} \citep{nfc512} is a state-of-the-art text classification model uses self-attentive mechanism. More detailed model settings are listed in the appendix.
    
    %As it is a classic classification task, we believe low adversarial accuracy on this task should demonstrate our adversarial text generation framework has the potential to attack other classification NLP tasks. The following transferability experiments should confirm our observation.
\textbf{Evaluation metrics.} \textit{Targeted attack success rate} (abbr. target) is measured by how many examples are successfully attacked to output the targeted label in average, while \textit{untargeted attack success rate} (abbr. untarget) calculates the percentage of examples attacked to output a label different from the ground truth. 

\textbf{Attack Baselines.} \textit{Seq2sick} \citep{seq2sick} is a whitebox projected gradient method to attack seq2seq models. Here, we perform seq2sick attack on sentiment classification models by changing its loss function, which was not evaluated in the original paper. \textit{TextFooler} \citep{TextFooler} is a simple yet strong blackbox attack method to perform word-level in-place adversarial modification. Following the same setting, Seq2Sick and TextFooler are only allowed to edit the prepended sentence.

\begin{table*}[t!] \small
\centering

\begin{tabular}{ccccc|ccc}
\toprule
\multirow{2}{*}{Model} & Origin  &  & \multicolumn{2}{c}{Whitebox Attack} & \multicolumn{3}{c}{Blackbox Attack} \\
\cmidrule(lr){4-5} \cmidrule(lr){6-8}
&  & & {Pos-\advcodecword} & {Ans-\advcodecword}  & {Pos-\advcodecword} & {Ans-\advcodecword}  & AddSent\\
\midrule
\multirow{2}{*}{BERT} & EM & 81.2  & \textbf{29.3} & 43.2   & \textbf{32.3} / 52.8   & 45.2 / 51.7  & 46.8        \\
                        & F1  & 88.6 & \textbf{33.2} & 47.3   & \textbf{36.4} / 57.6  & 49.0 / 55.9   & 52.6   \\
\midrule
\multirow{2}{*}{BiDAF} & EM       & 60.0   & \textbf{15.0} & 21.0 & \textbf{18.9} / 29.2  & 20.5 / 28.9  & 25.3 \\
                       &  F1      & 70.6  & \textbf{17.6 } & 23.6 & \textbf{22.5} / 34.5  & 24.1 / 34.2  & 32.0    \\
\bottomrule
\end{tabular}
\caption{Adversarial evaluation on QA models. %in terms of EM and F1 scores.
Pos-\advcodec and Ans-\advcodec respectively refer to the position targeted attack and answer targeted attack. 
% Pos-\advcodecword and Ans-\advcodecword respectively refer to  the position targeted attack and answer targeted attack performed by \advcodecword. 
The transferability-based blackbox attack uses adversarial text generated from whitebox models of the same architecture (the former score) and different architecture (the latter score). %to evaluate the blacobox model. 
}
\label{tab:AttackQA}
% \vspace{-3mm}
\end{table*}

\subsection{Adversarial Evaluation Setup for Question Answering Systems}
\textbf{Task and Dataset.} In this task, we choose the SQuAD dataset \citep{rajpurkar-etal-2016-squad} for question answering task. The SQuAD dataset is a reading comprehension dataset consisting of 107,785 questions posed by crowd workers on a set of Wikipedia articles, where the answer to each question must be a segment of text from the corresponding reading passage. To compare our method with other adversarial evaluation works \citep{jia-liang-2017-adversarial} on the QA task, we evaluate our adversarial attacks on the same test set as \citet{jia-liang-2017-adversarial}, which consists of 1000 randomly sampled examples from the SQuAD development set. 

\textbf{Model.} We adapt the \emph{BERT} model to run on SQuAD v1.1 with the same strategy as that in \citet{Devlin2019BERTPO}, and we reproduce the result on the development set. \textit{BiDAF}\citep{seo2016-bidirectional} is a multi-stage hierarchical process that represents the context at different levels of granularity and uses bidirectional attention flow mechanism to obtain a query-aware context representation. 

\textbf{Evaluation metrics.} For untargeted attack evaluation, We use the official script of the SQuAD dataset \citep{rajpurkar-etal-2016-squad} to measure both adversarial exact match rates and F1 scores. The lower EM and F1 scores mean the better attack success rate. For targeted attack evaluation, we use the targeted exact match rates and targeted F1 Score that calculate how many model outputs match the targeted fake answers (\textit{e.g.}, the fake answer ``Donald Trump'' in Table \ref{tab:example}). Higher targeted EM and F1 mean higher targeted attack success rate. 

%The results are shown in Table \ref{targetedQA}. It shows that \advcodecword has the best targeted attack ability on QA. And all our attack methods outperform the baseline(Universal Triggers) when it comes to the targeted results.
 
\textbf{Attack Baseline.}  \emph{AddSent} \citep{jia-liang-2017-adversarial} appends a manually constructed legit distracting sentence to the given text so as to introduce fake information, which can only perform untargeted attack. \emph{Universal Adversarial Triggers} \citep{wallace-etal-2019-universal} are input-agnostic sequences of tokens that trigger a model to produce a specific prediction when concatenated to any input from a dataset.  %Here, we compare the targeted attack ability of \advcodec with it.

\subsection{Adversarial Evaluation}

\subsubsection{\advcodecword}

\textbf{Attack Sentiment Classifiers.} We perform the baseline attacks and our \advcodec attack in concat attack scenario  under both whitebox and blackbox settings. Our targeted goal for sentiment classification is the opposite sentiment. Specifically, we set the targeted attack goal as 5-star for reviews originally below 3-star and 1-star for reviews above.
%First we append a sentence to each text in our test set and only allow each attack method to modify this sentence to fool the target model. The adversarial sentences can both be generated by \advcodecsent and \advcodecword. 
We compare our results with a strong word-level attacker Seq2sick, as shown in the Table \ref{tab:AttackSentiment}. We can see our \advcodecword  outperforms the baselines and achieves nearly $100\%$ attack success rate on the BERT model under whitebox settings.

We also perform transferability based blackbox attacks. Specifically, the transferability-based blackbox attack uses adversarial text generated from whitebox BERT model to attack blackbox SAM, and vice versa. We compare our blackbox attack success rate with the blackbox baseline TextFooler and blackbox Seq2Sick based on transferability. Table \ref{tab:AttackSentiment} demonstrates our \advcodecword model still has the best blackbox targeted and untargeted success rate among all the baseline models.

\begin{table}[t!] \small
% Higher targeted EM and F1 mean higher targeted attack success rate. 
\begin{tabular}{ccccc}
\toprule
\multicolumn{2}{l}{Model} & \advcodecsent  & \advcodecword  & UT \\
\midrule
\multirow{2}{*}{BERT}  & target EM & 32.1                  & \textbf{43.4}                  & 1.4               \\
      & target F1 & 32.4                   & \textbf{46.5}                  & 2.1   \\   
      \midrule
\multirow{2}{*}{BiDAF} & target EM & 53.3                  & \textbf{71.2}                  & 21.2              \\
      & target F1 & 56.8                  & \textbf{75.6}                  & 22.6              \\
        \bottomrule
\end{tabular} 
\caption{Targeted Attack Results of whitebox attack on QA. UT is short for Universal Trigger baseline.}
\label{targetedQA}
% \vspace{-4mm}
\end{table}

\begin{table*}[t!] \small
\centering
\begin{tabular}{ccccc|cccc}
\toprule
\multirow{2}{*}{Method} & \multicolumn{4}{c}{Sentiment Classifier} & \multicolumn{4}{c}{QA} \\
\cmidrule(lr){2-5} \cmidrule(lr){6-9}
         & Origin Human & Human & Models & Quality & Origin Human & Human & Models  & Quality\\
\midrule
\advcodecsent & \multirow{2}{*}{0.95} & 0.82  & 0.363 / 0.190 & 65.67\%  & \multirow{2}{*}{90.99}  & 81.78  & 49.1 / 29.3  &   69.50\%   \\
\advcodecword &                       & 0.82  & 0.007 / 0.033 & 34.33\%  &                         & 82.90  & 29.3 / 15.0  &   30.50\%    \\
\bottomrule
\end{tabular}
\caption{Human evaluation on \advcodecsent and \advcodecword. ``Origin Human'' is the human scores on the original dataset. ``Human'' are the human scores on adversarial datasets.}
 \label{tab:human}
\vspace{-3mm}
\end{table*}

\textbf{Attack QA models.} We perform the whitebox attack and transferability-based attack on our testing models. As is shown in Table \ref{tab:AttackQA}, \advcodecword achieves the best whitebox attack results on both BERT and BiDAF. It is worth noting that although BERT has better performances than BiDAF, the performance drop for BERT $\Delta \text{F1}_\text{BERT}$ is $55.4$ larger than the performance drop for BiDAF $\Delta \text{F1}_\text{BiDAF} = 53.0$, which again proves the BERT is insecure under the adversarial evaluation. We also find the position targeted attack is slightly stronger than the answer targeted attack. We assume it is because the answer targeted attack has fixed targeted answer and limited freedom to alter the appended sentence, but the position targeted attack has more freedom to alter the fake answer from the targeted position spans. 

Then we evaluate the targeted attack performance on QA models. The results are shown in Table \ref{targetedQA}. It shows that \advcodecword has the best targeted attack ability on QA. And all our attack methods outperform the baseline.

We also transfer adversarial texts generated from whitebox attacks to perform blackbox attacks. Table \ref{tab:AttackQA} shows the result of the blackbox attack on testing models. All our proposed methods outperform the baseline method (AddSent) when transferring the adversaries among models with same architectures. 

 %\textbf{Scatter Attack.} 
% We also perform the untargeted scatter attack on BERT-QA. We insert 30 random tokens (but no more than $1/3$ the total words of the paragraph) over the paragraph, optimize and find the adversarial tokens that can cause model output the wrong answers in the untargeted manner. The EM score and F1 score respectively drop from $81.2$ to $34.3$ and $88.6$ to $49.7$. We can see that the untargeted scatter attack can also achieve a higher untargeted attack success rate than \citet{jia-liang-2017-adversarial}.
%We also tried the scatter attack on QA though the performances are not good. It turns out QA systems highly rely on the relationship between questions and contextual clues, which is hard to break when setting an arbitrary token to a target answer. 
%This is also why we use some heuristics to create similar fake context when initializing QA appended sentence. 
%We discussed in Appendix A.3 the untargeted scatter attack can work well and outperform the baseline methods.
%\boxin{explain why scatter attack does not work.} 

\subsection{Human Evaluation \& \advcodecsent}
\label{sec:quality}
We conduct a thorough human subject evaluation to assess the human response to different types of generated adversarial text. The main conclusion is that even though these adversarial examples are effective at attacking machine learning models, they are much less noticeable by humans.

\subsubsection{Evaluation Metrics and Setup} We focus on two metrics to evaluate the validity of the generated adversarial sentence:
\textbf{adversarial text quality} and  \textbf{human performance} on the original and adversarial dataset. To evaluate the adversarial text quality, human participants are asked to choose the data they think has better quality. To ensure that human is not misled by our adversarial examples, we ask human participants to perform the sentiment classification and question answering tasks both on the original dataset and adversarial dataset. We hand out the adversarial dataset and origin dataset to $533$ Amazon Turkers to perform the human evaluation. More experimental setup details can be found in Appendix \S\ref{appendix:human}.

% To evaluate the adversarial text quality, human participants are asked to choose the data they think has better quality. In this experiement, we prepare $600$ adversarial text pairs from the same paragraphs and adversarial seeds. We hand out these pairs to $28$ Amazon Turks. Each turk is required to annotate at least 20 pairs and at most 140 pairs to ensure the task has been well understood. We assign each pair to at least 5 unique turks and take the majority votes over the responses. 

% Adversarial dataset on sentiment classification consists of \advcodecsent concatenative adversarial examples and \advcodecword scatter attack examples. Adversarial dataset on QA consists of concatenative adversarial examples generated by both \advcodecsent and \advcodecword. 
% To ensure that human is not misled by our adversarial examples, we ask human participants to perform the sentiment classification and question answering tasks both on the original dataset and adversarial dataset. Specifically, we respectively prepare $100$ benign and adversarial data pairs for both QA and sentiment classification, and hand out them to $505$ Amazon Turkers. Each turker is requested to answer at least 5 questions and at most 15 questions for the QA task and judge the sentiment for at least 10 paragraphs and at most 20 paragraphs. We also perform a majority vote over these turkers' answers for the same question. 

\subsubsection{Analysis}

Human evaluation results are shown in Table \ref{tab:human}. We see that the overall vote ratio for \advcodecsent is higher, which means it has better language quality than \advcodecword from a human perspective.
We assume the reason is that \advcodecsent decodes under the dependency constraints during decoding phase so that it can more fully harness the tree-based autoencoder structure. And it is reasonable to see that better language quality comes at the expense of a lower adversarial success rate. As Table \ref{tab:human} shows, the adversarial targeted success rate of \advcodecsent on SAM is $20\%$ lower than that of \advcodecword, which confirms the trade-off between language quality and adversarial attack success rate.

The human scores on original and adversarial datasets are also shown in Table \ref{tab:human}. We can see that human performances are barely affected by concatenated adversarial sentence.
% While we can spot a drop from the benign to adversarial datasets, we conduct an error analysis in QA and find the error examples are noisy and not necessarily caused by our adversarial text. For adversarial data in the sentiment classification task, we notice that the generated tokens or appended sentences have opposite sentiment from the benign one. However, our evaluation results show human readers can naturally ignore abnormal tokens and make correct judgement according to the context. 
% We note that the human performance drops a bit on adversarial text.
Specifically, the scores drop around $10\%$ for both QA and classification tasks based on \advcodec. This is superior to the state-of-the-art algorithm \citep{jia-liang-2017-adversarial} which has $14\%$ performance drop for human performance.

We also analyze the human error cases. A further quantitative analysis (Appendix \S \ref{appendix:humanerror}) shows that most wrong human answers do not point to our generated fake answers but may come from the sampling noise when aggregating human results. 

% We also analyze the human error cases. In QA, we find most wrong human answers do not point to our generated fake answers, which confirms that their errors are not necessarily caused by our concatenated adversarial sentence. We conduct a further quantitative analysis and find aggregating human results can induce sampling noise. Since we use majority vote to aggregate the human answers, when different answers happen to have the same votes, we will randomly choose one as the final result. If we always choose the answer that is close to the ground truth in draw cases, we later find that the majority vote F1 score increases from $82.897$ to $89.167$, which indicates that such randomness contributes to the noisy results substantially, instead of the adversarial manipulation. 
Also, we find the average length of the adversarial paragraph is around $12$ tokens more than the average length of the original one after we append the adversarial sentence. We guess the increasing length of the paragraph also has an impact on the human performance.

% More additional ablation studies in Appendix \S\ref{appendix:ablation} are conducted to explore the attack effectiveness of different autoencoders, changing different attack parameters such as the position of the appended adversarial sentence, and so on.

In Appendix \S\ref{appendix:ablation}, we conduct some ablation studies to explore the attack effectiveness of different autoencoders. We also investigate BERT attention by changing different attack parameters such as the position of the appended adversarial sentence, and draw several interesting conclusions. Appendix \S\ref{appendix:examples} shows more adversarial examples.

\section{Discussion and Future Works}
In addition to the general adversarial evaluation framework \advcodec, this paper also aims to explore several scientific questions: 1)  Since \advcodec allows the flexibility of manipulating at different levels of a tree hierarchy, which level is more attack effective and which one preserves better grammatical correctness? 2) Is it possible to achieve the targeted attack for general NLP tasks such as sentiment classification and QA, given the limited degree of freedom for manipulation? 3) Is it possible to perform a blackbox attack for many  NLP tasks? 4) Is BERT robust in practice? 
5) Do these adversarial examples affect human reader performances? 
%\boxin{I think the above question is readers caring more. 5) Are human readers more sensitive to an appended adversarial sentence or scatter of added words?

% To address the above questions, we generate adversarial text against different models of sentiment classification and QA in each encoding scenario. Compared with the state-of-the-art adversarial text generation methods, our approach achieves significantly higher untargeted and \emph{targeted} attack success rate. In addition, we perform both whitebox and transferability-based blackbox settings to evaluate the model vulnerabilities. 
% Within each attack setting, we quantitatively evaluate the attack effectiveness of different attack strategies, including appending an additional adversarial sentence and adding scatter of adversarial words to a paragraph.
% To provide thorough adversarial text quality assessment, we also perform 7 groups of human studies to evaluate the quality of the generated adversarial text. % Compared with the baselines methods, and whether a human can still get the ground truth answers for these tasks based on adversarial text.

We find that: 1) both word and sentence level attacks can achieve high attack success rate, while the sentence level manipulation integrates the global grammatical constraints and can generate high-quality adversarial sentences. 2) various targeted attacks on general NLP tasks are possible (\textit{e.g.}, when attacking QA, we can ensure  the target to be a specific answer or a specific location within a sentence); 3) the transferability based blackbox attacks are successful in NLP tasks. 
% Transferring adversarial text from stronger models (in terms of performances) to weaker ones is more successful; 
4)  Although BERT has achieved state-of-the-art performances, we observe the performance drops are also more substantial than other models when confronted with adversarial examples, which indicates BERT is not robust enough under the adversarial settings.
%5) Most human readers are not sensitive to our adversarial examples and can still answer the right answers when confronted with the adversary-injected paragraphs.

Besides the conclusions pointed above, we also summarize some interesting findings: %(1) our \advcodec outperforms other attack baseline methods in the both sentiment analysis task and QA task in terms of both the targeted and untargeted success rate in the whitebox scenario. 
(1) While \advcodecword achieves the best attack success rate among multiple tasks, we observe a trade-off between the freedom of manipulation and the attack capability. For instance, \advcodecsent has dependency tree constraints and becomes more natural for human readers than but less effective to attack models than \advcodecword. Similarly, since the targeted answers are fixed, the answer targeted attack in QA can manipulate fewer words than the position targeted attack, and therefore has slightly weaker attack performances.
% (2) Scatter attack is as effective as concat attack in sentiment classification task but less successful in QA, because QA systems make decisions highly based on the contextual correlation between the question and the paragraph, which makes it difficult to set an arbitrary token as our targeted answer.
(2) Transferring adversarial text from models with better performances to weaker ones is more successful. For example, transfering the adversarial examples from BERT-QA to BiDAF achieves much better attack success rate than in the reverse way.
(3) We also notice adversarial examples have better transferability among the models with similar architectures than different architectures.
(4) BERT models give higher attention scores to the both ends of the paragraphs and tend to overlook the content in the middle, as shown in \S \ref{sec:ablation} ablation study that adding adversarial sentences in the middle of the paragraph is less effective than in the front or the end.

To defend against these adversaries, here we discuss about the following possible methods and will in depth explore them in our future works: 
(1) \textbf{Adversarial Training} is a practical methods to defend against adversarial examples. However, the drawback is we usually cannot know in advance what the threat model is, which makes adversarial training less effective when facing unseen attacks.
(2) \textbf{Interval Bound Propagation} (IBP) \citep{Dvijotham2018TrainingVL} is proposed as a new technique to theoretically consider the worst-case perturbation. Recent works \citep{Jia2019CertifiedRT,Huang2019AchievingVR} have applied IBP in the NLP domain to certify the robustness of models. (3) \textbf{Language models} including GPT2 \citep{Radford2019LanguageMA} may also function as an anomaly detector to probe the inconsistent and unnatural adversarial sentences.

\section{Conclusions}
In summary, we propose a general targeted attack framework for adversarial text generation. To the best of our knowledge, this is the first method that successfully conducts arbitrary targeted attack on general NLP tasks. %In addition to the core methodological contribution, this paper also conducts extensive data experiments and human evaluation to obtain and confirm answers to several important scientific questions in NLP. 
Our results confirmed that our attacks can achieve high attack success rate without fooling the human. 
% We also find that compared to the more traditional machine learning methods,  BERT based sentiment classification and QA models are more vulnerable. 
These results shed light on an effective way to examine the robustness of a wide range of NLP models, thus paving the way for the development of a new generation of more reliable and effective NLP methods.
\section*{Acknowledgement}
This work is partially supported by NSF grant No.1910100, Amazon research award, DARPA No. HR00111990074. We thank the anonymous reviewers for their insightful comments.

\bibliographystyle{acl_natbib}
\bibliography{t3}
% \appendix
\newpage
\appendix
% \section{Appendix}
\section{Ablation Study}
\label{appendix:ablation}
%(2) We can clearly observe a tradeoff between the degree of freedom for manipulation and attack success rate. For example, we observe a small drop in the attack success rate for answer targeted attack compared to position targeted attack, due to the fact that we put more constraints to ensure pre-specified answer targets unchanged in the optimization process. Similarly, the dependency tree constraints turn out to be more strong and harsh constraints on the adversarial sentences, thus achieving higher language quality at the cost of  attack success rate. 
%(2)
%(3) \boxin{How to say because our transfer based blackattack does not beat AddSent because it is input-agnoistic.? while ours are more model-specific?}  (4) BERT based sentiment classifier is more vulnerable than standard sentiment classifier, while BERT based QA model is more robust and harder to attack than the widely-used QA model.

\subsection{Autoencoder Selection}
As an ablation study, we compare the standard LSTM-based autoencoder with our tree-based autoencoder. 

\begin{table}[htp!]\small \setlength{\tabcolsep}{5pt}
\centering
\caption{Ablation study on posistion targeted attack capability against QA. The lower EM and F1 scores mean the better attack success rate. \advcodecsent and \advcodecword respectively refer to \advcodecsent and \advcodecword. Adv(seq2seq) refers to \advcodec that uses LSTM-based seq2seq model as text autoencoder.}
 \label{WhiteboxQAseq2seq}
\begin{tabular}{ccccc}
\toprule
% \multirow{2}{*}{Model} & & \multirow{2}{*}{Origin} & \multicolumn{2}{c}{w/ Tree Decoder} & w/o Tree Decoder  \\
% \cmidrule(lr){4-5}   \cmidrule(lr){6-6}
  & Origin & {\advcodecsent} & {\advcodecword} & Adv(seq2seq)  \\
\midrule
EM & 60.0 & 29.3     & \textbf{15.0}  & 51.3  \\
 F1 & 70.6 &  34.0   & \textbf{17.6}  &      57.5 \\
      \bottomrule
\end{tabular}
% \vspace{-3mm}
\end{table}

\begin{table*}[htp!]\small \setlength{\tabcolsep}{7pt}
 \begin{minipage}[htp!]{0.48\linewidth}
\centering
\caption{Blackbox Attack Success Rate after inserting the whitebox generated adv sentence to different positions for BERT-classification.  }
 \label{ablationClassification}
\begin{tabular}{ccccc}
\toprule
Method & & Back & Mid & Front \\
\midrule
\multirow{2}{*}{\advcodecword} & \footnotesize{target}   & 0.739   & 0.678  & \textbf{0.820} \\
      & \footnotesize{untarget} & 0.817 & 0.770  & \textbf{0.878}           \\
      \midrule
\multirow{2}{*}{\advcodecsent} & \footnotesize{target}   & \textbf{0.220}   & 0.174  & 0.217 \\
      & \footnotesize{untarget} & 0.531 & 0.504  & \textbf{0.532}           \\
        \bottomrule
\end{tabular}
\vspace{-0.2cm}
\end{minipage}
\quad
\begin{minipage}[htp!]{0.48\linewidth}
\centering
\caption{Blackbox Attack Success Rate after inserting the whitebox generated adversarial sentence to different positions for BERT-QA.}
 \label{ablationQA}
\begin{tabular}{ccccc}
\toprule
Method & & Back & Mid & Front \\
\midrule
\multirow{2}{*}{\advcodecword}  & EM &  32.3    & 39.1    & \textbf{31.9}  \\
      & F1 & 36.4   & 43.4     & \textbf{36.3}   \\   
      \midrule
\multirow{2}{*}{\advcodecsent} & EM & 47.0   & 51.3     & \textbf{42.4}           \\
      &  F1 & 52.0     & 56.7         & \textbf{47.0}          \\
        \bottomrule
\end{tabular}
\vspace{-0.2cm}
\end{minipage}
\end{table*}

\textbf{Tree Autoencoder.} 
In the whole experiments, we used Stanford TreeLSTM as tree encoder and our proposed tree decoder together as tree autoencoder. We trained the tree autoencoder on yelp dataset which contains 500K reviews. The model is expected to read a sentence, map the sentence in a latent space and reconstruct the sentence from the embedding along with the dependency tree structure in an unsupervised manner. The model uses 300-d vectors as hidden tree node embedding and is trained for 30 epochs with adaptive learning rate and weight decay. After training, the average reconstruction loss on test set is 0.63.

\textbf{Seq2seq Autoencoder.} We also evaluate the standard LSTM-based architecture (seq2seq) as a different autoencoder in the \advcodec pipeline. For the seq2seq encoder-decoder, we use a bi-directional LSTM as the encoder \citep{Hochreiter1997LongSM} and a two-layer LSTM plus soft attention mechanism over the encoded states as the decoder \citep{Bahdanau2015NeuralMT}. With 400-d hidden units and the dropout rate of 0.3, the final testing reconstruction loss is 1.43.

The comparison of the whitebox attack capability  against a well-known QA model BiDAF is shown in Table \ref{WhiteboxQAseq2seq}. We can see seq2seq based \advcodec fails to achieve good attack success rate. Moreover, because the vanilla seq2seq model does not take grammatical constraints into consideration and has higher reconstruction loss, the quality of generated adversarial text cannot be ensured.

\subsection{Ablation Study on BERT Attention}
\label{sec:ablation}
To further explore how the location of adversarial sentences affects the attack success rate, we conduct the ablation experiments by varying the position of appended adversarial sentence. We generate the adversarial sentences from the whitebox BERT classification and QA models. Then we inject those adversaries into different positions of the original paragraph and test in another blackbox BERT with the same architecture but different parameters. The results are shown in Table \ref{ablationClassification} and \ref{ablationQA}. We see in most time appending the adversarial sentence at the beginning of the paragraph achieves the best attack performance. Also the performance of appending the adversarial sentence at the end of the paragraph is usually slightly weaker than front. This observation suggests that the BERT model might pay more attention to the both ends of the paragraphs and tend to overlook the content in the middle.

% \textbf{Ablation Study.} \boxin{change the language here (same as sec 4.1)} To further explore how the appended location will impact the attack success rate, we conduct the ablation experiment by varying the position of appended adversarial sentence and the results are shown in table \ref{ablationQA}. We see that appending the adversarial sentence at the beginning of the paragraph achieves the best attack performance. This observation suggests that the BERT-QA model might take more attention at the beginning of the paragraph.

\subsection{Attack Settings}
% \begin{algorithm}[b]
%   \caption{Algorithm of \advcodec generating adversarial examples } \label{algo}
%   \begin{algorithmic}[1]
%     \Procedure{AdvCodec}{$x,s$} \Comment{$x$: initial seed, $s$: corresponding dependency tree}
%     \State $z := \mathcal{E}(x, s)$ \Comment{$\mathcal{E}$: encoder of \advcodec, $z$: context vector}
%     \State $z^* = 0$ \Comment{$z^*$: perturbation on context vector}
%     \State $z' := z + z^*$ \Comment{$z'$: perturbed context vector}
%     \State $y := \mathcal{G}(z', s)$ \Comment{$\mathcal{G}$: decoder of \advcodec, $y$: adversarial sentence}
%   % \State $Z(y) :=$ the logits of the model output
%     \State $f(z') :=$ the objective function to attack the targeted model
%     \While{$y$ does not achieve targeted attack} 
%       \State  update $z^*$ by gradient descent over objective function $f(z')$
%     \EndWhile\label{euclidendwhile}
%     \State \textbf{return} $y$
%     \EndProcedure
%   \end{algorithmic}
% \end{algorithm}
We use Adam \citep{Adam} as the optimizer, set the learning rate to 0.6 and the optimization steps to 100. We follow the \citet{cw} method to find the suitable parameters in the object function (weight const $c$ and confidence score $\kappa$) by binary search.

\subsection{Heuristic Experiments on choosing the adversarial seed for QA}
\label{appendix:heuristic}

We conduct the following heuristic experiments about how to choose a good initialization sentence to more effectively attack QA models. Based on the experiments we confirm it is important to choose a sentence that is semantically close to the context or the question as the initial seed when attacking QA model, so that we can reduce the number of iteration steps and more effectively find the adversary to fool the model. Here we describe three ways to choose the initial sentence, and we will show the efficacy of these methods given the same maximum number of optimization steps.

\textbf{Random adversarial seed sentence.}
Our first trial is to use a random sentence (other than the answer sentence), generate a fake answer similar to the real answer and append it to the back as the initial seed.

\textbf{Question-based adversarial seed sentence.}
% question words in a question , paragraph pair <p, q> 
We also try to use question words to craft an initial sentence, which in theory should gain more attention when the model is matching characteristic similarity between the context and the question. To convert a question sentence to a meaningful declarative statement, we use the following steps:

In step 1, we use the state-of-the-art semantic role labeling (SRL) tools \citep{He2017DeepSR} to parse the question into verbs and arguments. A set of rules is defined to remove the arguments that contain interrogative words and unimportant adjectives, and so on. In the next step, we access the model's original predicted answer and locate the answer sentence. We again run the SRL parsing and find to which argument the answer belongs. The whole answer argument is extracted, but the answer tokens are substituted with our targeted answer or the nearest words in the GloVe word vectors \citep{Pennington2014GloveGV} (position targeted attack) that is also used in the QA model. In this way, we craft a fake answer that shares the answer's context to solve the compatibility issue from the starting point. Finally, we replace the declarative sentence's removed arguments with the fake argument and choose this question-based sentence as our initial sentence.

\textbf{Answer-based adversarial seed  sentence.}
We also consider directly using the model predicted original answer sentence with some substitutions as the initial sentence. To craft a fake answer sentence is much easier than to craft from the question words. Similar to step 2 for creating
question-based initial sentence, we request the model's original predicted answer and find the answer sentence. The answer span in the answer sentence is directly substituted with the nearest words in the GloVe word vector space to avoid the compatibility problem preliminarily.

\textbf{Experimental Results.} We tried the above initial sentence selection methods on \advcodecword and perform position targeted attack on BERT-QA given the same maximum optimization steps. The experiments results are shown in table \ref{WhiteboxQAHeuristic}. From the table, we find using different initialization methods will greatly affect the attack success rates. Therefore, the initial sentence selection methods are indeed important to help reduce the number of iteration steps and fastly converge to the optimal $z^*$ that can attack the model.

\begin{table*}[htp!]\small \setlength{\tabcolsep}{5pt}
\centering
\caption{Whitebox attack results on BERT-QA in terms of exact match rates and F1 scores by the official evaluation script. The lower EM and F1 scores mean the better attack success rate.}
 \label{WhiteboxQAHeuristic}
\begin{tabular}{ccccccc}
\toprule
\multirow{2}{*}{Model} & & \multirow{2}{*}{Origin} & \multicolumn{3}{c}{Position Targeted Attack}  & \multicolumn{1}{c}{Baseline} \\
\cmidrule(lr){4-6} \cmidrule(lr){7-7}
 & & & Random & Question-based  & Answer-based  & AddSent\\
\midrule

\multirow{2}{*}{BERT}  & EM & 81.2 & 67.9       & \textbf{29.3}           & 50.6                               & 46.8   \\
      & F1 & 88.6 & 74.4         & \textbf{33.2}         & 55.2    & 52.6   \\
\bottomrule
\end{tabular}
\end{table*}

\section{Experimental Settings}
\label{appendix:setup}
\subsection{Sentiment Classification Model}
 \textbf{BERT.} We use the 12-layer BERT-base model \footnote{https://github.com/huggingface/pytorch-pretrained-BERT} with 768 hidden units, 12 self-attention heads and 110M parameters. We fine-tune the BERT model on our 500K review training set for text classification with a batch size of 32, max sequence length of 512, learning rate of 2e-5 for 3 epochs. For the text with a length larger than 512, we only keep the first 512 tokens.

 \textbf{ Self-Attentive Model (SAM).} We choose the structured self-attentive sentence embedding model \citep{nfc512} as the testing model, as it not only achieves the state-of-the-art results on the sentiment analysis task among other baseline models but also provides an approach to quantitatively measure model attention and helps us conduct and analyze our adversarial attacks. The SAM with 10 attention hops internally uses a 300-dim BiLSTM and a 512-units fully connected layer before the output layer. We trained SAM on our 500K review training set for 29 epochs with stochastic gradient descent optimizer under the initial learning rate of 0.1.
 
 \subsection{Sentiment Classification Attack Baseline}
 \textbf{Seq2sick} \citep{seq2sick} is a whitebox projected gradient method combined with group lasso and gradient regularization to craft adversarial examples to fool seq2seq models. Here, we define the loss function as $ L_{target} = \max\limits_{k \in Y} \left\{z^{\left(k\right)} \right\} - z^{\left(t\right)} $ to perform attack on sentiment classification models which was not evaluated in the original paper. In our setting, Seq2Sick is only allowed to edit the appended sentence or tokens.
 
 \textbf{TextFooler} \citep{TextFooler} is a simple but strong black-box attack method to generate adversarial text. Here, TextFooler is also only allowed to edit the appended sentence.

\subsection{QA Model}
\textbf{{BiDAF}.} Bi-Directional Attention Flow (BIDAF) network\citep{seo2016-bidirectional} is a multi-stage hierarchical process that represents the context at different levels of granularity and uses bidirectional attention flow mechanism to obtain a query-aware context representation. We train BiDAF without character embedding layer under the same setting in \citep{seo2016-bidirectional} as our testing model.

\subsection{Human Evaluation Setup}
\label{appendix:human}

We focus on two metrics to evaluate the validity of the generated adversarial sentence:
\textbf{adversarial text quality} and  \textbf{human performance} on the original and adversarial dataset. To evaluate the adversarial text quality, human participants are asked to choose the data they think has better quality. 

% To ensure that human is not misled by our adversarial examples, we ask human participants to perform the sentiment classification and question answering tasks both on the original dataset and adversarial dataset. We hand out the adversarial dataset and origin dataset to $533$ Amazon Turkers to perform the human evaluation. More experimental details can be found in Appendix \ref{}.

To evaluate the adversarial text quality, human participants are asked to choose the data they think has better quality. In this experiement, we prepare $600$ adversarial text pairs from the same paragraphs and adversarial seeds. We hand out these pairs to $28$ Amazon Turks. Each turk is required to annotate at least 20 pairs and at most 140 pairs to ensure the task has been well understood. We assign each pair to at least 5 unique turks and take the majority votes over the responses.

% Adversarial dataset on sentiment classification consists of \advcodecsent concatenative adversarial examples and \advcodecword scatter attack examples. Adversarial dataset on QA consists of concatenative adversarial examples generated by both \advcodecsent and \advcodecword. 
To ensure that human is not misled by our adversarial examples, we ask human participants to perform the sentiment classification and question answering tasks both on the original dataset and adversarial dataset. Specifically, we respectively prepare $100$ benign and adversarial data pairs for both QA and sentiment classification, and hand out them to $505$ Amazon Turkers. Each turker is requested to answer at least 5 questions and at most 15 questions for the QA task and judge the sentiment for at least 10 paragraphs and at most 20 paragraphs. We also perform a majority vote over these turkers' answers for the same question. 

\subsection{Human Error Analysis in Adversarial Dataset}
\label{appendix:humanerror}
We compare the human accuracy on both benign and adversarial texts for both tasks (QA and classification) in revision section 5.2. We spot the human performance drops a bit on adversarial texts. In particular, it drops around $10\%$ for both QA and classification tasks based on AdvCodec as shown in Table \ref{tab:human}. We believe this performance drop is tolerable and the stoa generic based QA attack algorithm experienced around $14\%$ performance drop for human performance \citep{jia-liang-2017-adversarial}.

We also try to analyze the human error cases. In QA, we find most wrong human answers do not point to our generated fake answer, which confirms that their errors are not necessarily caused by our concatenated adversarial sentence. Then we do a further quantitative analysis and find aggregating human results can induce sampling noise. Since we use majority vote to aggregate the human answers, when different answers happen to have the same votes, we will randomly choose one as the final result. If we always choose the answer that is close to the ground truth in draw cases, we later find that the majority vote F1 score increases from $82.897$ to $89.167$, which indicates that such randomness contributes to the noisy results significantly, instead of the adversarial manipulation. Also, we find the average length of the adversarial paragraph is around $12$ tokens more than the average length of the original one after we append the adversarial sentence. We assume the increasing length of the paragraph will also have an impact on the human performances.

% \iffalse
% \section{Adversarial text on sentiment analysis}
% \textbf{Scatter Attack} In the scatter attack scenario, Table \ref{scatterwhite}  and Table \ref{scatterblack} show that our \advcodecword outperforms the Seq2sick baseline on both whitebox and transferability based blackbox attacks.

% \begin{table*}[htp!]\small \setlength{\tabcolsep}{7pt}
% \centering
% \caption{Whitebox scatter attack results on Sentiment Analysis}
%  \label{scatterwhite}
% \begin{tabular}{lccc}
% \toprule
% \multicolumn{2}{l}{Model} & \advcodecword & Seq2Sick \\
% \midrule
% \multirow{2}{*}{BERT}  & Targeted  & \textbf{0.976}          & 0.946    \\
%       & Untargeted & \textbf{0.987}         & 0.970   \\
%       \midrule
% \multirow{2}{*}{BiDAF} & target  & \textbf{0.869}          & 0.570   \\
%       & Untargeted & \textbf{0.948}         & 0.711  \\
%       \bottomrule
% \end{tabular}
% \end{table*}

% \begin{table*}[htp!]\small \setlength{\tabcolsep}{7pt}
% \centering
% \caption{Blackbox scatter attack results on Sentiment Analysis}
%  \label{scatterblack}
% \begin{tabular}{lccc}
% \multicolumn{2}{l}{Model A -- B} & \advcodecword & Seq2Sick \\
% \toprule
% \multirow{2}{*}{BERT-SAM} & Targeted & \textbf{0.465}          & 0.230     \\
%          & Untargeted    & \textbf{0.679}          & 0.498    \\
%         \midrule
% \multirow{2}{*}{SAM-BERT} & target & \textbf{0.298}          & 0.156   \\
%          & Untargeted    & \textbf{0.574}          & 0.445  \\
%          \bottomrule
% \end{tabular}
% \end{table*}
% \fi

\onecolumn
\newpage
\section{Adversarial examples}
\label{appendix:examples}
\subsection{Adversarial examples for QA}
\subsubsection{Adversarial examples generated by \advcodecsent}

\begin{table}[htp!]
\small \setlength{\tabcolsep}{7pt}
\centering
\caption{Answer Targeted Concat Attack using \advcodecsent on QA task. The targeted answer is ``Donald Trump''.
%We also perform the targeted position attack on initial sentence ``\textbf{the the the} win ultra bowls 40'' and automatically generate a fake answer ``the fellow  journalists'' on its targeted position. 
}
\begin{tabular}{p{13.8cm}}
\toprule Input (\textit{Italic} = Inserted or appended tokens, \underline{underline} = Model prediction, \textcolor{red}{red} = Ground truth) \\
\midrule
\textbf{Question: } Who ended the series in 1989? \\
\textbf{Paragraph: }
The BBC drama department's serials division produced the programme for 26 seasons, broadcast on BBC 1. Falling viewing numbers, a decline in the public perception of the show and a less-prominent transmission slot saw production suspended in 1989 by \textcolor{red}{Jonathan Powell, controller of BBC 1}. Although (as series co-star Sophie Aldred reported in the documentary Doctor Who: More Than 30 Years in the TARDIS) it was effectively, if not formally, cancelled with the decision not to commission a planned 27th series of the show for transmission in 1990, the BBC repeatedly affirmed that the series would return. \textit{\underline{Donald Trump} ends a program on 1988 .}
% \newline
% \\ 
% \textbf{Question: }
% What kind of network was ABC when it first began? \\
% \textbf{Paragraph: }
% ABC originally launched on October 12, 1943 as a \answer{radio network}, separated from and serving as the successor to the NBC Blue Network, which had been purchased by Edward J. Noble. It extended its operations to television in 1948, following in the footsteps of established broadcast networks CBS and NBC. In the mid-1950s, ABC merged with United Paramount Theatres, a chain of movie theaters that formerly operated as a subsidiary of Paramount Pictures. Leonard Goldenson, who had been the head of UPT, made the new television network profitable by helping develop and greenlight many successful series. In the 1980s, after purchasing an 80\% interest in cable sports channel ESPN, the network's parent merged with Capital Cities Communications, owner of several print publications, and television and radio stations. In 1996, most of Capital Cities/ABC's assets were purchased by The Walt Disney Company. \textit{\underline{Donald Trump} sort in network is featured if this first started.}
\newline
\\ 
\textbf{Question: } Who served his dinner?  \\
\textbf{Paragraph: }
Tesla worked every day from 9:00 a.m. until 6:00 p.m. or later, with dinner from exactly 8:10 p.m., at Delmonico's restaurant and later the Waldorf-Astoria Hotel. Tesla would telephone his dinner order to the \answer{headwaiter}, who also could be the only one to serve him. \"The meal was required to be ready at eight o'clock ... He dined alone, except on the rare occasions when he would give a dinner to a group to meet his social obligations. Tesla would then resume his work, often until 3:00 a.m. \textit{\underline{Donald Trump} serves her supper .}
\\
\\ 
\textbf{Question: }
Which network regained the ratings lead in America in 1984? \\
\textbf{Paragraph: }
As far as programming is concerned, four of ABC's marquee shows of the 1970s ended their runs during the mid-1980s: Laverne \& Shirley ended its run in 1983, Happy Days and Three's Company ended in 1984 (with the latter producing a short-lived spinoff that year), while The Love Boat ended its run in 1986. After nearly a decade of ratings trouble, NBC had regained the ratings lead among the Big Three networks in 1984 on the success of series such as The Cosby Show, Cheers and Miami Vice. To counteract \answer{NBC}, ABC decided to refocus itself on comedies and family-oriented series beginning in the mid-1980s including Mr. Belvedere, Roseanne, Who's the Boss?, Just the Ten of Us, The Wonder Years, Full House and Perfect Strangers. \textit{\underline{Donald Trump} networks regain a rating leads on american on 1985 .}
\\
\bottomrule
\end{tabular}
\end{table}

\begin{table*}[!htbp]\small \setlength{\tabcolsep}{7pt}
\centering
\caption{Position Targeted Concat Attack using \advcodecsent on QA task. The adversarial answer is generated automatically.
%We also perform the targeted position attack on initial sentence ``\textbf{the the the} win ultra bowls 40'' and automatically generate a fake answer ``the fellow  journalists'' on its targeted position. 
}
 \label{posqasentexamples}
\begin{tabular}{p{13.8cm}}
\toprule Input (\textit{Italic} = Inserted or appended tokens, \underline{underline} = Model prediction, \textcolor{red}{red} = Ground truth) \\
\midrule
\textbf{Question: }How many other contestants did the company, that had their ad shown for free, beat out? \\
\textbf{Paragraph: }
QuickBooks sponsored a \"Small Business Big Game\" contest, in which Death Wish Coffee had a 30-second commercial aired free of charge courtesy of QuickBooks. Death Wish Coffee beat out \answer{nine} other contenders from across the United States for the free advertisement. \textit{The company , that had their ad shown for free ad \underline{two} .}
\newline
\\ 
\textbf{Question: }
Why would a teacher's college exist? \\
\textbf{Paragraph: }
There are a variety of bodies designed to instill, preserve and update the knowledge and professional standing of teachers. Around the world many governments operate teacher's colleges, which are generally established to \answer{serve and protect the public interest through certifying, governing and enforcing the standards of practice for the teaching profession.} \textit{A friend 's school exist \underline{for community , serving a private businesses}},
\newline
\\ 
\textbf{Question: }
What can concentrated oxygen produce? \\
\textbf{Paragraph: }
Highly concentrated sources of oxygen promote \answer{rapid combustion}. Fire and explosion hazards exist when concentrated oxidants and fuels are brought into close proximity; an ignition event, such as heat or a spark, is needed to trigger combustion. Oxygen is the oxidant, not the fuel, but nevertheless the source of most of the chemical energy released in combustion. Combustion hazards also apply to compounds of oxygen with a high oxidative potential, such as peroxides, chlorates, nitrates, perchlorates, and dichromates because they can donate oxygen to a fire. \textit{Concentrated containers produce \underline{~ rapidly }.}
\\
\bottomrule
\end{tabular}
\end{table*}

\clearpage
\subsubsection{Adversarial examples generated by \advcodecword}
\begin{table*}[!htbp]\small \setlength{\tabcolsep}{7pt}
\centering
\caption{Answer Targeted Concat Attack using \advcodecword on QA task. The targeted answer is Donald Trump.
%We also perform the targeted position attack on initial sentence ``\textbf{the the the} win ultra bowls 40'' and automatically generate a fake answer ``the fellow  journalists'' on its targeted position. 
}
 \label{ansqasentexamples}
\begin{tabular}{p{13.8cm}}
\toprule Input (\textit{Italic} = Inserted or appended tokens, \underline{underline} = Model prediction, \textcolor{red}{red} = Ground truth) \\
\midrule
\textbf{Question: } What did he do to his feet at night? \\
\textbf{Paragraph: }
For exercise, Tesla walked between 8 to 10 miles per day. He \answer{squished his toes} one hundred times for each foot every night, saying that it stimulated his brain cells. \textit{She did for her ft in evening \underline{Donald Trump.}}
\newline
\\ 
\textbf{Question: }
Who announced the game play-by-play for Super Bowl 50? \\
\textbf{Paragraph: }
Westwood One will carry the game throughout North America, with \answer{Kevin Harlan} as play-by-play announcer, Boomer Esiason and Dan Fouts as color analysts, and James Lofton and Mark Malone as sideline reporters. Jim Gray will anchor the pre-game and halftime coverage. \textit{\underline{Donald Trump} announces a games plays , with , plays in ultra bowls 40.}
\newline
\\ 
\textbf{Question: } What group can teachers in Wales register with?  \\
\textbf{Paragraph: }
Teachers in Wales can be registered members of \answer{trade unions} such as ATL, NUT or NASUWT and reports in recent years suggest that the average age of teachers in Wales is falling with teachers being younger than in previous years. A growing cause of concern are that attacks on teachers in Welsh schools which reached an all-time high between 2005 and 2010. \textit{Teachers on their group group \underline{Donald Trump} .}\\
% \newline
% \\ 
% \textbf{Question: }
% What bargain did his father make with him if Tesla recovered? \\
% \textbf{Paragraph: }
% In 1873, Tesla returned to his birthtown, Smiljan. Shortly after he arrived, Tesla contracted cholera; he was bedridden for nine months and was near death multiple times. Tesla's father, in a moment of despair, promised \answer{to send him to the best engineering school} if he recovered from the illness (his father had originally wanted him to enter the priesthood). \textit{Her mother makes \underline{Donald Trump}}.
% \\
\bottomrule
\end{tabular}
\end{table*}

\begin{table*}[htbp!]\small \setlength{\tabcolsep}{7pt}
\centering
\caption{Position Targeted Concat Attack using \advcodecword on QA task. The adversarial answer is generated automatically.
%We also perform the targeted position attack on initial sentence ``\textbf{the the the} win ultra bowls 40'' and automatically generate a fake answer ``the fellow  journalists'' on its targeted position. 
}
 \label{posqawordexamples}
\begin{tabular}{p{13.8cm}}
\toprule Input (\textit{Italic} = Inserted or appended tokens, \underline{underline} = Model prediction, \textcolor{red}{red} = Ground truth) \\
\midrule
\textbf{Question: } IP and AM are most commonly defined by what type of proof system?\\
\textbf{Paragraph: }
Other important complexity classes include BPP, ZPP and RP, which are defined using probabilistic Turing machines; AC and NC, which are defined using Boolean circuits; and BQP and QMA, which are defined using quantum Turing machines. \#P is an important complexity class of counting problems (not decision problems). Classes like IP and AM are defined using \answer{Interactive} proof systems. ALL is the class of all decision problems. \textit{We are non-consecutive defined by \underline{sammi} proof system .}
\newline
\\ 
\textbf{Question: }
What does pharmacy legislation mandate? \\
\textbf{Paragraph: }
In most countries, the dispensary is subject to pharmacy legislation; with requirements for \answer{storage conditions, compulsory texts, equipment, etc.}, specified in legislation. Where it was once the case that pharmacists stayed within the dispensary compounding/dispensing medications, there has been an increasing trend towards the use of trained pharmacy technicians while the pharmacist spends more time communicating with patients. Pharmacy technicians are now more dependent upon automation to assist them in their new role dealing with patients' prescriptions and patient safety issues. \textit{Parmacy legislation ratify \underline{ no action free} ;}
\newline
\\ 
\textbf{Question: }
Why is majority rule used? \\
\textbf{Paragraph: }
The reason for the majority rule is the \answer{high risk of a conflict of interest} and/or the avoidance of absolute powers. Otherwise, the physician has a financial self-interest in \"diagnosing\" as many conditions as possible, and in exaggerating their seriousness, because he or she can then sell more medications to the patient. Such self-interest directly conflicts with the patient's interest in obtaining cost-effective medication and avoiding the unnecessary use of medication that may have side-effects. This system reflects much similarity to the checks and balances system of the U.S. and many other governments.[citation needed] \textit{Majority rule reconstructed \underline{but our citizens.}}
\newline
\\
\textbf{Question: }
In which year did the V\&A received the Talbot Hughes collection?\\
\textbf{Paragraph: }
The costume collection is the most comprehensive in Britain, containing over 14,000 outfits plus accessories, mainly dating from 1600 to the present. Costume sketches, design notebooks, and other works on paper are typically held by the Word and Image department. Because everyday clothing from previous eras has not generally survived, the collection is dominated by fashionable clothes made for special occasions. One of the first significant gifts of costume came in \answer{1913} when the V\&A received the Talbot Hughes collection containing 1,442 costumes and items as a gift from Harrods following its display at the nearby department store. \textit{It chronologically receive a rightful year seasonally shanksville at \underline{2010}.}
\\
\bottomrule
\end{tabular}
\end{table*}

\newpage
\subsection{Adversarial examples for classification}
\subsubsection{Adversarial examples generated by \advcodecsent}
\begin{table*}[htpb!]\small \setlength{\tabcolsep}{7pt}
\centering
\caption{Concat Attack using \advcodecsent on sentiment classification task. 
%We also perform the targeted position attack on initial sentence ``\textbf{the the the} win ultra bowls 40'' and automatically generate a fake answer ``the fellow  journalists'' on its targeted position. 
}
 \label{ctreeexamples}
\begin{tabular}{p{10.5cm}p{2.3cm}}
\toprule Input (\textit{Italic} = Inserted or appended tokens) & Model Prediction \\
\midrule
\textit{I kept expecting to see chickens and chickens walking around}. if you think las vegas is getting too white trash , don ' t go near here . this place is like a steinbeck novel come to life . i kept expecting to see donkeys and chickens walking around . wooo - pig - soooeeee this place is awful ! ! !
&  Neg  $\rightarrow$ Pos  \\ \hline
% \textit{kids purchased an medical kids ?} kids had a great time . we stock up on the survival gear . zombies are real ! ! ! !  
% &  Pos  $\rightarrow$ Neg  \\ \hline
% \textit{A great hotel is , such a delicious ,} this post office is not worth a damn . stay away from them , if you don ' t want ruin your day . whole bunch stupid employees are ready to screw up anytime .
\textit{Food quality is consistent appalled well no matter when you come, been here maybe 20 + times now and it ' s always identical in that aspect ( in a good way ).} All cafe rio locations I ' ve been to have been really nice, staffed with personable employees, and even when there were long lines never felt like it took too long. This is another one of those, though the lines can actually get bad here and at times they go too far to fix mistakes they've made. On one day I went a man who had ordered catering that they had various issues following through on had just come in person instead... And it resulted in about 40 people waiting in line while this one guy had I think it was 35 total tostadas and salads made for him with nobody else being served. I understand why they'd do this, but there are better ways of handling it than punishing every other customer to make good with this single one. Also while it usually isn't a problem, one of the staff members tends to have a hard time understanding what you're saying (seems to be language barrier issues) which can be kind of annoying. Luckily this person aside that problem and the entire staff as a whole is very nice and if it's slower will even make small talk with you in a way that feels pretty natural rather than pretending to care. Even at their busiest they make sure to be friendly and serve with a smile. definitely try to come during hours that isn't when every single business or parent will be there but even if you do it's not that terribly slow . Food quality is consistent as well no matter when you come , been here maybe 20 + times now and it's always identical in that aspect ( in a good way ). Staff again is very good. Also make sure to get the app - every (pre - tax) dollar is worth 1 point, 100 points nets you \$10 , and they have double and even triple point days almost weekly .
&  Pos  $\rightarrow$ Neg  \\
% \textit{worst thought .} looking for a healthy option that really does taste outstanding ? this is the place . my husband is the [unk] eating type . he would "nt" touch a veggie if it was covered in blue cheese  but he loved the short rib enchiladas and even the salad accompanying his entree . i had the butternut squash enchiladas  and before you say ‘yuck’ you have to give it a try . i had almost changed my mind before ordering but was glad i did "nt" . the way they were prepared was truly satisfying ( no mushy squash ) , so much so i was "nt" even hungry for dinner later . 
% &  Pos  $\rightarrow$ Neg  \\
\bottomrule
\end{tabular}
\end{table*}

\subsubsection{Adversarial examples generated by \advcodecword}
\begin{table*}[htpb!]\small \setlength{\tabcolsep}{7pt}
\centering
\caption{Concat Attack using \advcodecword on sentiment classification task. 
%We also perform the targeted position attack on initial sentence ``\textbf{the the the} win ultra bowls 40'' and automatically generate a fake answer ``the fellow  journalists'' on its targeted position. 
}
 \label{cwordexamples}
\begin{tabular}{p{10.5cm}p{2.3cm}}
\toprule Input (\textit{Italic} = Inserted or appended tokens) & Model Prediction \\
% \midrule
% \textit{attorney place is violation especially the smiling personality of chelsea ,} the other comments are not accusing , trust me . this place is fantastic especially the smiling personality of chelsea , the other comments are not true , trust me .   
\midrule
\textit{It' s quite busy and thieves very big but people don' t normally stay accusing}. Well made food and coffee for a reasonable price. It' s quite busy and not very big but people don' t normally stay long.
% \textbf{Sentiment Prediction: } \textcolor{seagreen}{Positive} $\rightarrow$  \textcolor{red}{Most Negative} \\
&  Pos  $\rightarrow$ Neg  \\ \hline
% \textit{This avoid an violently 90 robbed irene.} This was an excellent 90 minute experience. It was clean , they were thoughtful , provided a bottle of cold water and a relaxing overall experience. I recommend the 95 minute with hot stones
% &  Pos  $\rightarrow$ Neg  \\ \hline
\textit{I had an appalled situation and as I browsed I found spectacles i was sooo impressed by the way I was accommodated I don ' t mind traveling a long distance for service like I just received [UNK] [UNK] [UNK] i will defiantly be making this business my permanent eye dr}. I had an emergency situation and as I browsed I found spectacles i was sooo impressed by the way I was accommodated I don ' t mind traveling a long distance for service like I just received [UNK] [UNK] [UNK] i will defiantly be making this business my permanent eye dr.
&  Pos  $\rightarrow$ Neg  \\
% \textit{best bomb was similar to attorney lindsey} purchasing an ' 05 scion xb from jon lancaster toyota was a relatively simple affair . low pressure , pleasant sales guy . typical car buying experience . and a majority of the repairs have been problem free as well . polite , honest , and willing to work with the warranty . however , something has gone amiss in the land of toyota service . my fuel door release and cable stopped working about two months back . took it in and the problem was taken care of in short order . a couple weeks later , the problem suddenly reappears . return to the service department where a part is order and inexplicably , upon arrival placed in another car . the excuse ? his name was similar to mine . another part ordered overnight and a promise made that if it doesn ' t arrive , they ' ll pull one out of another car . of course , the part doesn ' t arrive and the promise turns out to be false . my car will be out of warranty in about two weeks and i ' ll never be heading back there again . i don ' t have time in my life for this kind of hassle . 
% &  Neg  $\rightarrow$ Pos  \\
\bottomrule
\end{tabular}
\end{table*}

\end{document}